\documentclass{article}
\usepackage{spconf,amsmath,graphicx,hyperref}
\usepackage{algorithm}
\usepackage{algorithmic}
\usepackage{arydshln}
\usepackage[table]{xcolor}
\usepackage{amssymb}
\usepackage{listings}
\usepackage{xcolor}
\usepackage{multirow}
\usepackage{booktabs}
\usepackage{threeparttable}
\usepackage{caption}
\usepackage{textcomp}
\usepackage{parskip}
\usepackage{enumitem}

\usepackage{authblk}

\title{WenetSpeech-Chuan: A Large-Scale Sichuanese Corpus with Rich Annotation for Dialectal Speech Processing}
\name{%
  {\fontsize{10.5pt}{12pt}\selectfont
  Yuhang Dai$^{1,*}$, Ziyu Zhang$^{1,*}$, Shuai Wang$^{4,5}$, Longhao Li$^{1}$, Zhao Guo$^{1}$, Tianlun Zuo$^{1}$, Shuiyuan Wang$^{1}$, Hongfei Xue$^{1}$, \\
  \textit{Chengyou Wang$^{1}$, Qing Wang$^{3}$, Xin Xu$^{2}$, Hui Bu$^{2}$, Jie Li$^{3}$, Jian Kang$^{3}$, Binbin Zhang$^{4}$, Lei Xie$^{1,\dagger}$}%
  }%
}
\address{%
  {\fontsize{10.5pt}{12pt}\selectfont
  $^1$\text{Audio, Speech and Language Processing Group (ASLP@NPU), School of Computer Science} \\ 
  Northwestern Polytechnical University \\
  $^2$\text{Beijing AISHELL Technology Co., Ltd.} \\
  $^3$\text{Institute of Artificial Intelligence (TeleAI), China Telecom} \\
  $^4$\text{School of Intelligence Science and Technology, Nanjing University} \\
  $^5$\text{WeNet Open Source Community}%
  }%
}

\begin{document}

\ninept
\twocolumn[
\begin{center}
  {{\fontsize{13pt}{14pt}\selectfont \bfseries WENETSPEECH-CHUAN: A LARGE-SCALE SICHUANESE CORPUS WITH RICH \\ 
  \vspace{0.5em}
   ANNOTATION FOR DIALECTAL SPEECH PROCESSING}}

  \vspace{0.5em}

  {\fontsize{10.5pt}{12pt}\selectfont
    \textit{Yuhang Dai$^{1,*}$, Ziyu Zhang$^{1,*}$, Shuai Wang$^{4,5}$, Longhao Li$^{1}$, Zhao Guo$^{1}$, Tianlun Zuo$^{1}$, Shuiyuan Wang$^{1}$, Hongfei Xue$^{1}$}, \\
    \textit{Chengyou Wang$^{1}$, Qing Wang$^{3}$, Xin Xu$^{2}$, Hui Bu$^{2}$, Jie Li$^{3}$, Jian Kang$^{3}$, Binbin Zhang$^{4}$, Lei Xie$^{1,\dagger}$}
  }

  \vspace{0.8em}

  {\fontsize{10.5pt}{12pt}\selectfont
  $^1$ Audio, Speech and Language Processing Group (ASLP@NPU), School of Computer Science, \\ 
  Northwestern Polytechnical University \\
  $^2$ Beijing AISHELL Technology Co., Ltd. \\
  $^3$ Institute of Artificial Intelligence (TeleAI), China Telecom \\
  $^4$ School of Intelligence Science and Technology, Nanjing University \\
  $^5$ WeNet Open Source Community
  }
  
  \vspace{1em}
\end{center}
]

\begin{abstract}
\vspace{0.3cm}
The scarcity of large-scale, open-source data for dialects severely hinders progress in speech technology, a challenge particularly acute for the widely spoken Sichuanese dialects of Chinese. To address this critical gap, we introduce \textbf{WenetSpeech-Chuan}, a 10,000-hour, richly annotated corpus constructed using our novel \textbf{Chuan-Pipeline}, a complete data processing framework for dialectal speech. To facilitate rigorous evaluation and demonstrate the corpus's effectiveness, we also release high-quality ASR and TTS benchmarks, \textbf{WenetSpeech-Chuan-Eval}, with manually verified transcriptions. Experiments show that models trained on WenetSpeech-Chuan achieve state-of-the-art performance among open-source systems and demonstrate results comparable to commercial services. As the largest open-source corpus for Sichuanese dialects, WenetSpeech-Chuan not only lowers the barrier to research in dialectal speech processing but also plays a crucial role in promoting AI equity and mitigating bias in speech technologies. The corpus, benchmarks, models, and receipts are publicly available on our \href{https://github.com/ASLP-lab/WenetSpeech-Chuan}{project page}.
\end{abstract}


\section{Introduction}
\label{sec:intro}
In recent years, large-scale open-source datasets have substantially accelerated progress in automatic speech recognition (ASR) and text-to-speech (TTS) tasks. However, both tasks still face significant challenges when applied to accented and dialectal speech. Prior studies~\cite{dialect01, Ddialect_aierland, dialect_Vietnamese} have demonstrated that ASR systems often struggle with dialects due to pronunciation differences and acoustic mismatches. Notably, performance drops are observed even for speech with only slight accents~\cite{AccentFold}. Similarly, research on accented TTS~\cite{accent_tts, accent_tts02} has highlighted the difficulty of modeling accurate accent variability.

This challenge is particularly acute within the diverse landscape of Chinese dialects. The Sichuan-Chongqing dialects (Sichuanese dialects~\footnote{\url{https://en.wikipedia.org/wiki/Sichuanese_dialects}}), spoken by approximately 120 million people, constitutes one of the largest dialect communities in Southwest China. Its tonal system, vocabulary, and grammar differ substantially from Standard Mandarin, creating clear linguistic distinctions. Consequently, the scarcity of dialect-specific data severely degrades the performance of mainstream ASR and TTS systems for Sichuanese speakers, highlighting a critical need for a large-scale, open-source corpus dedicated to the Sichuanese dialects. Despite this clear need, existing open-source resources for Sichuanese dialects are critically insufficient in both scale and diversity.

The only publicly available datasets are two small corpora from MagicData, which we refer to as: MagicData onversation~\footnote{\url{https://magichub.com/datasets/sichuan-dialect-conversational-speech-corpus/}} with 4.53 hours of speech and MagicData Daily-Use~\footnote{\url{https://magichub.com/datasets/sichuan-dialect-scripted-speech-corpus-daily-use-sentence/}} with 6.4 hours of speech. While KeSpeech~\cite{kespeech} also contains a small subset of Southwestern Mandarin accents, these samples represent accented Mandarin rather than dialectal speech. Due to their limited scale and narrow coverage, these resources are inadequate for training robust ASR and TTS systems for Sichuanese dialects.

Drawing on our experience in building large-scale corpora with the WenetSpeech series~\cite{wenetspeech, li2025wenetspeech, ma2024wenetspeech4tts}, we now address the critical resource gap in Sichuanese dialectal speech by introducing WenetSpeech-Chuan. This large-scale, richly annotated\footnote{transcription, domain labels, and multiple paralinguistic annotation such as age, gender and emotion} corpus features over 10,000 hours of Sichuanese dialects speech, sourced from diverse domains such as short videos, entertainment, and live streams. To enable the construction of this corpus and to facilitate future research, we also propose Chuan-Pipeline, a comprehensive data processing framework designed specifically for creating high-quality speech resources from raw Sichuanese dialects data. Finally, to ensure rigorous and reproducible evaluation, we release two benchmark sets, WSC-Eval-ASR and WSC-Eval-TTS, which include fine-grained manual corrections. Our contributions can be summarized as follows:

\begin{itemize}[itemsep=2pt, topsep=2pt, parsep=0pt, partopsep=0pt]
\item We release WenetSpeech-Chuan, the largest-to-date open-source corpus for the Sichuanese dialects, containing over 10,000 hours of richly annotated speech data from diverse, real-world domains.
\item We develop and open-source Chuan-Pipeline, a systematic data processing framework that enables the creation of high-quality dialectal speech datasets.
\item We establish strong ASR and TTS benchmarks on our new evaluation sets, demonstrating that models trained on WenetSpeech-Chuan achieve state-of-the-art performance in both recognition accuracy and synthesis quality for the Sichuanese dialects.
\end{itemize}

\begin{figure*}[t] 
  \centering
  \vspace{-0.5cm}
  \includegraphics[clip, trim=1cm 8cm 1cm 8cm, width=\textwidth]{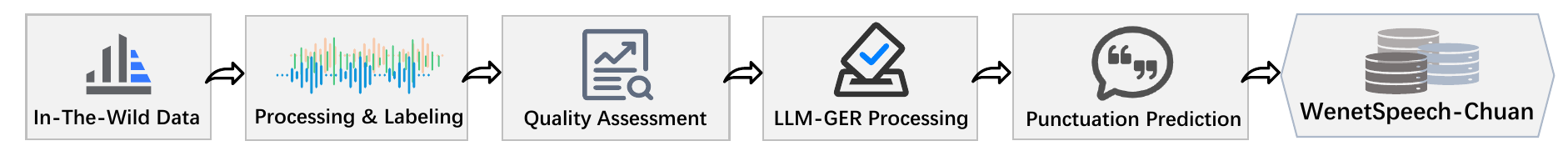}
  \caption{Overview of the WenetSpeech–Chuan pipeline}
  \label{fig:chuan_pipe}
  \vspace{-0.3cm}
\end{figure*}

\vspace{-0.3cm}
\section{Chuan-Pipeline}
To construct the WenetSpeech-Chuan dataset,
we propose a comprehensive speech data processing pipeline, \textbf{Chuan-Pipeline}, as illustrated in Fig.~\ref{fig:chuan_pipe}. This pipeline systematically transforms raw, unlabeled audio into a richly annotated corpus suitable for advanced ASR and TTS research.

\subsection{Pre-Processing and Labeling}
The initial stage of the pipeline focuses on data acquisition, segmentation, and the enrichment of speech segments with multi-dimensional paralinguistic labels. Raw data acquisition begins with mining metadata from online video platforms to identify content potentially containing Sichuanese dialects. Following an initial manual verification to confirm the presence of the target dialect, the acquired audio streams undergo a multi-stage workflow:

\begin{itemize}[leftmargin=10pt]
\item \textbf{VAD \& Segmentation:} Long audio streams are segmented into 5-25 second clips using Voice Activity Detection (VAD), removing non-speech portions like silence and noise.
\item \textbf{Single-Speaker Selection \& Clustering:} We first employ the pyannote toolkit\footnote{https://github.com/pyannote} to isolate single-speaker segments. Subsequently, speaker embeddings are extracted with the CAM++ model~\cite{wang2023cam++} and clustered to assign a consistent speaker ID to all utterances from a single individual.
\item \textbf{Paralinguistic Annotation:} Speaker gender is identified using a pre-trained classifier\footnote{https://github.com/JaesungHuh/voice-gender-classifier} (98.7\% accuracy). Age is estimated via the Vox-Profile benchmark\footnote{https://github.com/tiantiaf0627/vox-profile-release} and categorized into age stages(children, teenager, young, middle-aged, old). Emotion is labeled by a majority vote over predictions from Emotion2vec~\cite{ma2023emotion2vec} and SenseVoice\footnote{https://github.com/plexormedia/sensevoice}, covering seven categories (happy, angry, sad, neutral, fearful, surprised, and disgusted).
\end{itemize}

\subsection{Quality Assessment}
To ensure the audio quality of processed data, we implement an automated quality assessment stage. To select data across different quality levels, we use timestamp-aligned speech as input and extract metrics such as duration and Signal-to-Noise Ratio (SNR). These features are then used to compute a Word-level Virtual Mean Opinion Score (WVMOS), which serves as a proxy for perceptual audio quality. Low-quality audio samples are then discarded.

\subsection{LLM-GER Processing}

\begin{figure}[ht]
    \centering
    \includegraphics[clip, trim=5cm 4cm 3cm 5cm, width=1.2\linewidth]{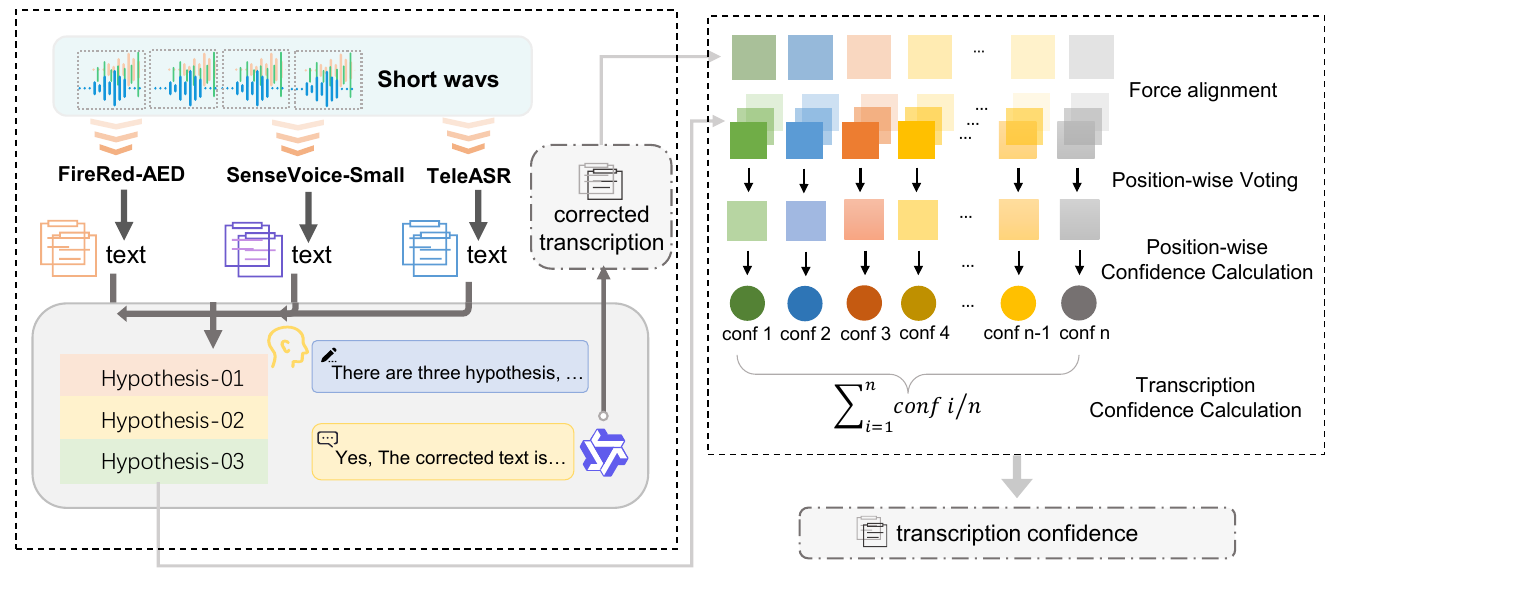}  
    \caption{LLM Generative Error Correction based Rover (LLM-GER) process for Sichuanese dialects.}
    \captionsetup[figure]{skip=2pt} 
    \label{fig:GER}
    \vspace{-0.3cm}
\end{figure}

To enhance the accuracy of automatic speech recognition (ASR) transcriptions, building upon prior research~\cite{GER_MU}, we propose a robust ASR transcription framework tailored for the Sichuanese dialects. Our approach, termed \textbf{LLM} \textbf{G}enerative \textbf{E}rror Correction based \textbf{R}OVER (\textbf{LLM-GER}), aims to merge outputs from multiple ASR systems into a single accurate and reliable transcription. First, three different ASR systems (FireRed-ASR, SenseVoice-Small, and TeleASR) produce initial candidate transcriptions. These are then merged by Qwen3~\cite{yang2025qwen3}, which leverages its strong dialectal understanding and our carefully designed prompt to perform error correction without altering the original semantics or token length. The prompt will be publicly available on our \href{https://github.com/ASLP-lab/WenetSpeech-Chuan}{project page}. Finally, transcription confidence is calculated based on the four transcriptions, as detailed in Figure~\ref{fig:GER}.

Through this approach, we fully leverage the ability of LLMs to normalize Sichuanese dialectal expressions while integrating the complementary strengths of multiple ASR systems. This combination produces high-quality transcriptions for WenetSpeech-Chuan. Calculations on the test set show that LLM-GER improves transcription accuracy by approximately 15\% on average compared to individual ASR transcription.

\subsection{Punctuation Prediction}
Accurate transcriptions with punctuation are crucial for TTS training, yet text-only punctuation prediction often misaligns with actual speech pauses. To address this, we propose a multi-modal punctuation prediction method combining audio and text modality. For audio modality, we use Kaldi model\footnote{https://github.com/kaldi-asr/kaldi} to force-align audio with text, obtaining the word timestamps and pause durations, classifying pauses into short or long pauses based on thresholds (e.g., 0.25s for short, 0.5s for long). For text, a BiLSTM punctuation model predicts marks at pause candidates: \textbf{commas} for short pauses and \textbf{periods}, \textbf{question marks}, or \textbf{exclamations} for long pauses. Thresholds are iteratively refined through human feedback to ensure punctuation aligns with actual speech pauses.

\begin{figure}[!htbp]
    \centering
    \vspace{-0.35cm}
    \includegraphics[clip, trim=9.5cm 8cm 10cm 7.5cm, width=1\linewidth]{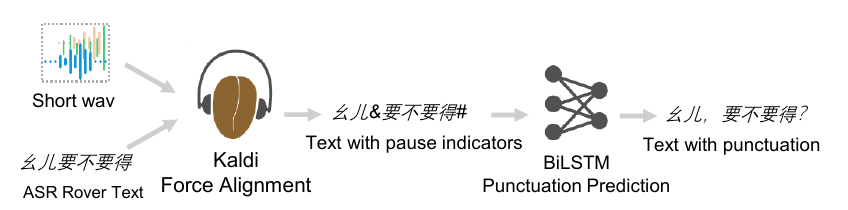} 
    \caption{Multimodal punctuation prediction process for Sichuanese dialects}
    \captionsetup[figure]{skip=5pt} 
    \label{fig:placeholder}
    \vspace{-0.6cm}
\end{figure}

\section{The WenetSpeech-Chuan Corpus}

By applying the Chuan-Pipeline to the collected multi-source raw data, we have constructed the \textbf{WenetSpeech-Chuan} corpus, a large-scale, multi-label, and multi-domain resource for the Sichuanese dialects. In this section, we describe the corpus in detail, covering its metadata, audio format, data diversity, as well as the design principles of the training and evaluation sets.

\subsection{Data Size and Confidence}
We assign confidence for each segment of audio that measures the quality of the ASR transcription. As shown in Table \ref{ChuanSpeech_partition}, we select 3,714 hours Strong Label data, whose confidence is greater than 0.90. The 6,299 hours Weak Label data, whose confidence is between 0.60 and 0.90, is reserved in our metadata for semi-supervised or other usage. In summary, WenetSpeech-Chuan has 10,013 hours of raw audio.

\begin{table}[!t]
    \centering
    \vspace{-0.2cm}
    \caption{WenetSpeech-Chuan Corpus Partition}
    \vspace{-0.4cm}
    \label{ChuanSpeech_partition}
    \setlength{\tabcolsep}{14pt} 
    \renewcommand{\arraystretch}{1} 
    \begin{tabular}{ccc}
        \hline 
        Set          & Confidence & Duration (h)      \\ \hline 
        Strong Label &  [0.9, 1.0] &   3,714    \\
        Weak Label   &  [0.6, 0.9) &   6,299    \\ \hline 
        Total   &       /     &   10,013    \\ \hline 
    \end{tabular}
    \vspace{-0.7cm}
\end{table}

\subsection{Domain Distribution} 
\begin{figure}[ht]
    \centering
    \includegraphics[clip, trim=7cm 4cm 6cm 4cm, width=1.1\linewidth]{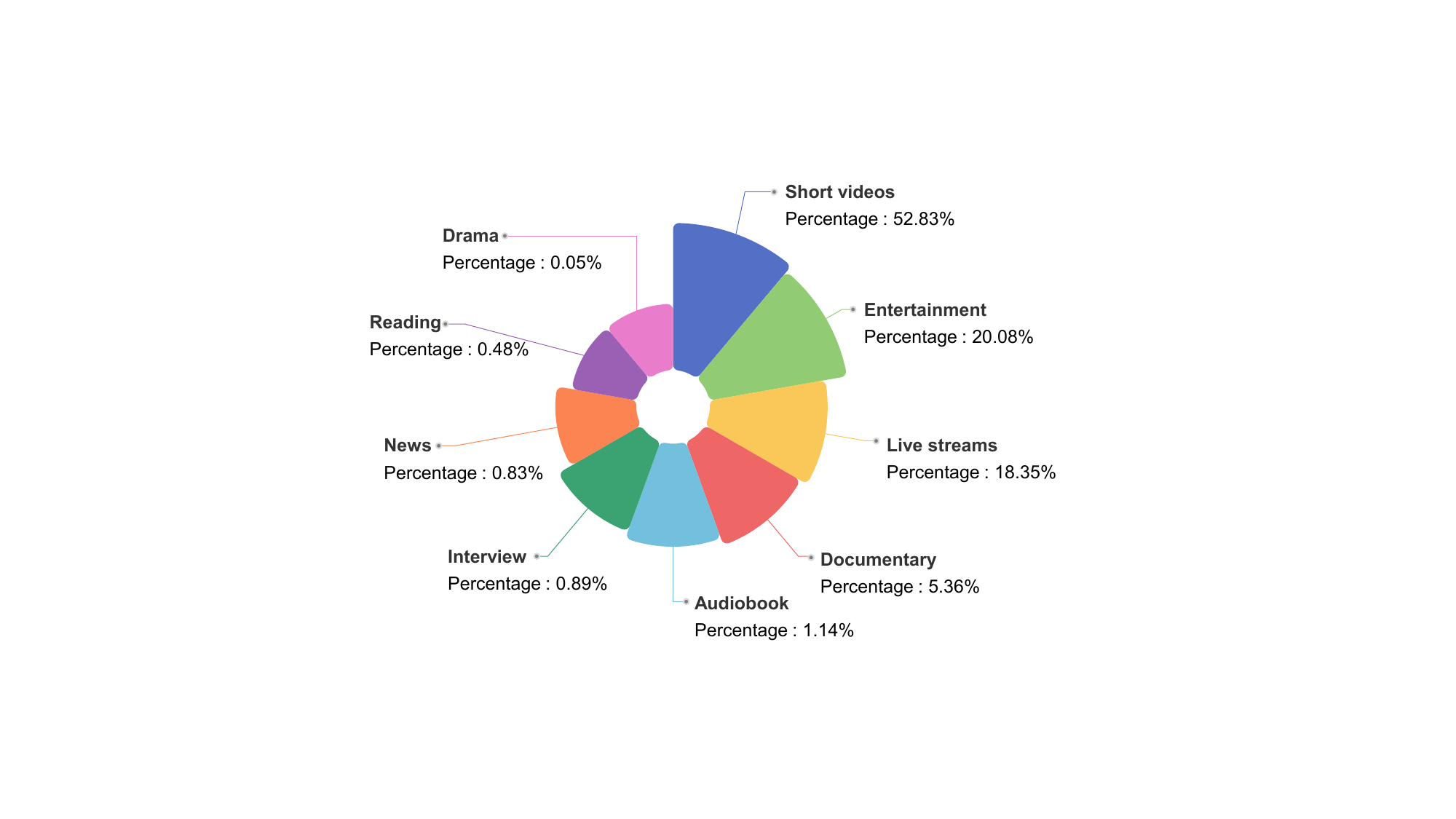}  
    \caption{Domain distribution of WenetSpeech-Chuan dataset.}
    \captionsetup[figure]{skip=2pt} 
    \label{fig:domain}
    \vspace{-0.4cm}
\end{figure}

We summarize the source domains of WenetSpeech-Chuan in Fig.~\ref{fig:domain}, which consists of 9 categories. Short videos account for the largest share (52.83\%), followed by entertainment (20.08\%) and live streams (18.35\%). Other domains, including documentaries, audiobooks, interviews, news, reading, and drama, make up smaller proportions but enhance the dataset’s diversity.

\subsection{Quality Distribution}
As illustrated in Fig.~\ref{fig:placeholder}, the audio quality scores, derived from our WVMOS-based metric, are concentrated in the 2.5 to 4.0 range, with a significant peak between 3.0 and 3.5. This distribution indicates that the majority of the corpus consists of high-to-moderate quality speech, striking a balance between clean recordings and real-world acoustic conditions, thereby making it robust for training general-purpose speech models.
\begin{figure}[ht]
    \centering
    \vspace{-0.3cm} 
    \includegraphics[width=0.85\linewidth]
    {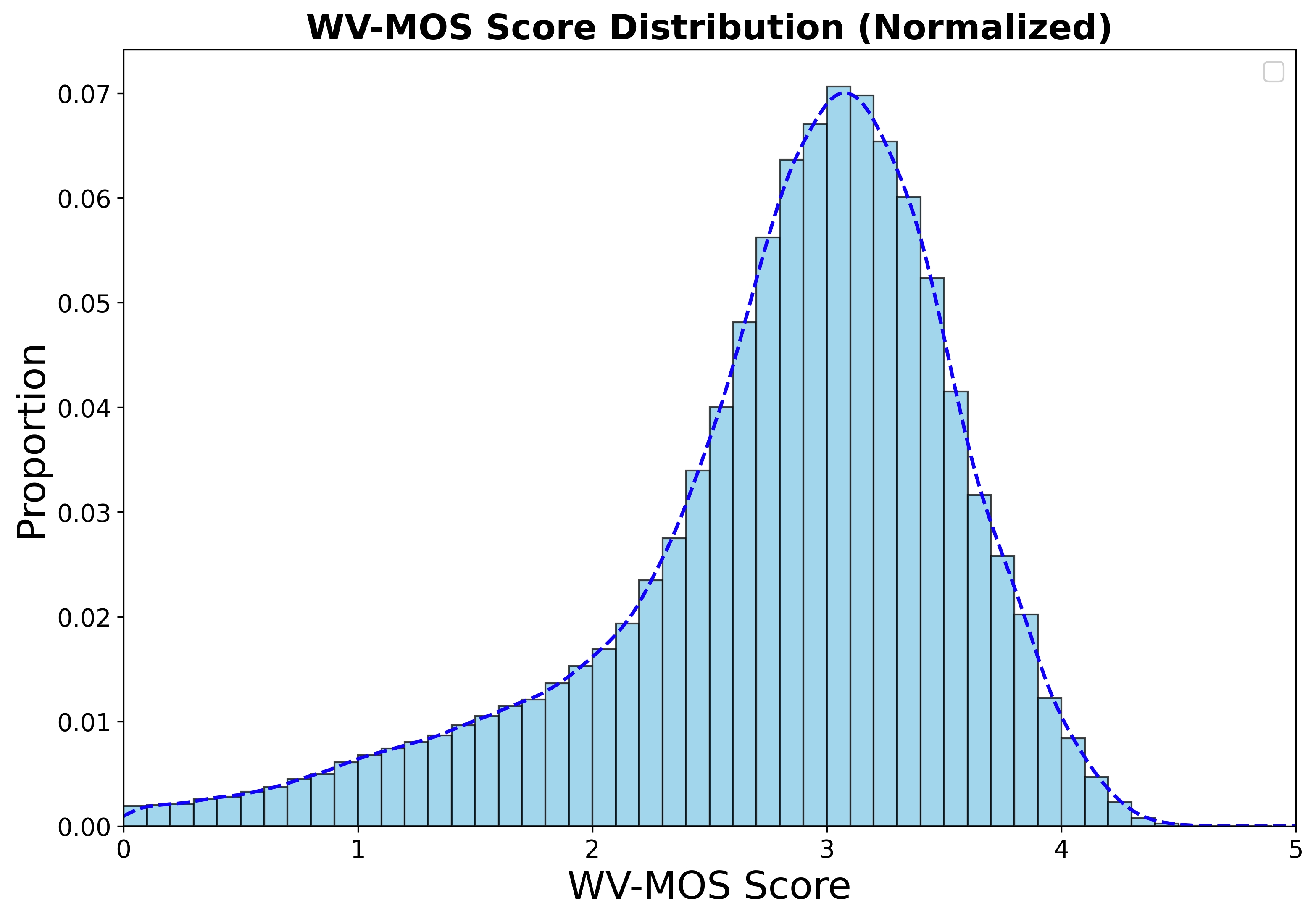} 
    \caption{Quality distribution of WenetSpeech-Chuan dataset.}
    \label{fig:placeholder}
    \vspace{-0.4cm}
\end{figure}


\subsection{WenetSpeech-Chuan Eval Benchmark}
\textbf{ASR Evaluation Set} To address the absence of a standardized benchmark for Sichuanese dialects ASR, we constructed \textit{WSC-Eval-ASR}. We first preprocessed raw Sichuanese dialects data from various domains using the Chuan-pipeline, and then manually refined by professional human annotators. Furthermore, all audio samples are annotated with speaker attributes, including age, gender, and emotional state. To facilitate fine-grained performance analysis, we further partitioned this 9.7-hour set into Easy and Hard subsets based on the source domain and acoustic environment. This allows for a more nuanced evaluation of model robustness. Detailed statistics are provided in Table~\ref{ChuanSpeech_Eval}.

\begin{table}[t]
    \centering
    \vspace{-0.2cm}
    \caption{Details of WSC-Eval-ASR Set.}
    \vspace{-0.4cm}
    \label{ChuanSpeech_Eval}
    \setlength{\tabcolsep}{2.5pt} 
    \renewcommand{\arraystretch}{1} 
    \begin{tabular}{cccc}
        \hline 
        Set &  Main Domain & Duration (h)   \\ \hline
        Easy    & audio book, reading         & 8.55        \\
        Hard    & short videos, entertainment, drama        & 1.15       \\ \hline 
        Total      & /       & 9.7            \\ \hline 
    \end{tabular}
    \vspace{-0.8cm}
\end{table}

\textbf{TTS Evaluation Set}  To address the absence of Sichuanese dialects TTS test sets, we construct \textit{WSC-Eval-TTS} two subsets: \textit{WSC-Eval-TTS-easy}, containing sentences with dialectal words across diverse domains, and \textit{WSC-Eval-TTS-hard}, consisting of long sentences and LLM–generated sentences in varied styles such as tongue twisters, folk sayings and emotional speech. For audio prompts, we select 10 speakers (5 male and 5 female) from MagicData and internal recordings, each recording 200 sentences to ensure balance across gender, age, and accent variations.

\begin{table*}[ht]
    \centering
    \small
    \caption{ASR results (CER\%$\downarrow$) on Sichuanese datasets. \textbf{Bold} indicates best performance, \underline{underlined} indicates second-best performance, and \colorbox{green!8}{light green background} indicates models finetuned on a high-quality internal corpus to show the system' potential as a foundation model.}
    \vspace{-0.4cm}
    \setlength{\tabcolsep}{3pt} 
    \renewcommand{\arraystretch}{1.1} 
    \begin{tabular}{lc@{\hspace{5pt}}c@{\hspace{15pt}}c@{\hspace{15pt}}c@{\hspace{5pt}}ccc}
        \toprule
        \multirow{2}{*}{\textbf{Model}} &
          \multirow{2}{*}{\textbf{Model Size}} &
          \multicolumn{3}{c}{\textbf{WSC-Eval-ASR}} &
          \multicolumn{2}{c}{\textbf{Magicdata}}  &
          \multirow{2}{*}{\textbf{Avg.}} \\\cmidrule(lr){3-5} \cmidrule(lr){6-7}
         &
          \textbf{} &
          \textbf{Easy} &
          \textbf{Hard} &
          \textbf{Total} &
          \textbf{Conversation} &
          \textbf{Daily-Use} &
          \textbf{} \\
          \midrule
        \textbf{with LLM} & & & & & & & \\  
        Kimi-Audio~\cite{kimiaudio} & 7B  & 16.65  & 28.66  & 17.66 & 24.67 & \textbf{5.77}  & 18.68 \\
        FireRedASR-LLM~\cite{fireredasr}   & 8.3B & {12.80}  & {25.27} & {14.40}  & {17.68} & {6.69} & {15.37} \\
        Qwen2.5-omni~\cite{Qwen2.5-Omni}         & 3B  & 16.94  & 26.01  & 18.20  & 20.40 & 6.32  & 17.69 \\
        Qwen2.5-omni-WSC-Finetune  & 3B  & {14.36}  & {24.14}  & {15.61} & {18.45} & 6.15  & {15.74} \\
        \rowcolor{green!8} Qwen2.5-omni+internal data  & 3B  & {13.17}  & {23.36}  & {14.81} & {18.50} & 5.88  & {15.14} \\
        \rowcolor{green!8}  Qwen2.5-omni-WSC-Finetune + internal data  & 3B  & {12.93}  & {23.19}  & {14.25} & {17.95} & \underline{5.89}  & {14.84} \\
        \midrule
        \textbf{without LLM} & & & & & & & \\  
        SenseVoice-small~\cite{sensevoice}  & 234M & 17.43 & 28.38  & 18.39 & 23.50 & 8.77  & 19.29 \\
        Whisper~\cite{whisper}      & 244M & 52.06 & 63.99  & 53.59 & 55.88 & 52.03  & 55.51 \\
        FireRedASR-AED~\cite{fireredasr}   & 1.1B & 13.29 & 23.64  & 14.62 & 17.84 & 6.69  & 15.14 \\
        Paraformer~\cite{paraformer}      & 220M & 14.34 & 24.61  & 15.66 & 19.81 & 8.16  & 16.52 \\
         Paraformer-WSC-Finetune    & 220M & {12.15} & {22.6}  & {13.51} & {16.60} & 8.02 & {14.58} \\
        \rowcolor{green!8} Paraformer + internal data    & 220M & \underline{11.93}  & \underline{21.82}  & \underline{13.14} & \underline{15.61} & 6.77 & \underline{13.85} \\
        \rowcolor{green!8} Paraformer-WSC-Finetune + internal data   & 220M & \textbf{11.59} & \textbf{21.59}  & \textbf{12.87} & \textbf{14.59} & {6.28} & \textbf{13.38} \\
        \bottomrule
    \end{tabular}
    \label{tab:eval_results}
\end{table*}

\begin{table*}[ht]
\vspace{-0.2cm}
\centering
\small
\caption{TTS results on WSC-Eval-TTS set. \textbf{Bold} indicates best performance, \underline{underlined} indicates second-best performance, and \colorbox{green!8}{light green background} indicates models trained on WenetSpeech-Chuan or additionally finetuned on an internal high-quality dataset.}
\vspace{-0.4cm}
\setlength{\tabcolsep}{1pt}
\renewcommand{\arraystretch}{1.1}

\begin{threeparttable}
\begin{tabular}{lcccccccccc}
\toprule
\multirow{2}{*}{\textbf{Model}} & \multicolumn{5}{c}{\textbf{WSC-Eval-TTS-easy}} & \multicolumn{5}{c}{\textbf{WSC-Eval-TTS-hard}} \\ 
\cmidrule(lr){2-6} \cmidrule(lr){7-11}
& \textbf{CER(\%)$\downarrow$} & \textbf{SIM(\%)$\uparrow$} & \textbf{IMOS$\uparrow$} & \textbf{SMOS$\uparrow$} & \textbf{AMOS$\uparrow$} & \textbf{CER(\%)$\downarrow$} & \textbf{SIM(\%)$\uparrow$} & \textbf{IMOS$\uparrow$} & \textbf{SMOS$\uparrow$} & \textbf{AMOS$\uparrow$} \\ 
\midrule
Step-Audio-TTS~\cite{huang2025step}    & 10.83 & 67.66 & 3.81 & 2.86 & 3.15 & 12.52 & 54.52 & 3.75 & 2.77 & 3.06 \\
CosyVoice 2.0~\cite{du2024cosyvoice}   & 7.14  & 70.27 & 3.88 & 3.10 & 3.69 & 9.06  & 60.10 & 3.96 & 2.73 & 3.81 \\
Qwen-TTS\tnote{$\dagger$}              & \underline{4.13} & - & 3.95 & - & 3.90 & \underline{7.35} & - & \textbf{4.02} & - & 3.88 \\
\rowcolor{green!8}
CosyVoice2-WSC                      & 4.28  & \underline{72.78} & \textbf{4.13} & \underline{3.94} & \underline{4.05} & 8.78  & \underline{62.59} & 3.85 & \underline{2.78}  & \underline{3.92} \\
\rowcolor{green!8}
CosyVoice2-WSC-SFT  & \textbf{4.08} & \textbf{78.84} & \underline{4.10} & \textbf{4.16} & \textbf{4.20} & \textbf{7.22} & \textbf{67.96} &  \underline{4.01}  &  \textbf{3.03}  &  \textbf{3.98}  \\
\bottomrule
\end{tabular}

\begin{tablenotes}
\footnotesize
\item[$\dagger$] Commercial system with single fixed speaker, and speaker similarity is not considered.
\end{tablenotes}
\end{threeparttable}
\label{tab:comparison}
\vspace{-0.6cm}
\end{table*}

\vspace{-0.2cm}
\section{Experiments}

\subsection{Automatic Speech Recognition}

To validate the effectiveness of the proposed dataset, we evaluated different models on three distinct test sets: our proposed \textit{WSC-Eval-ASR}, and two public datasets, MagicData-Conversation and MagicData-Daily-Use. The benchmark included a suite of off-the-shelf systems, from dedicated ASR models like SenseVoice~\cite{sensevoice}, Whisper-small~\cite{whisper}, and Paraformer~\cite{paraformer}, to omni-modal LLMs such as Kimi-Audio~\cite{kimiaudio} and Qwen2.5-omni. To directly demonstrate the value of our corpus, we fine-tuned both Paraformer and Qwen2.5-Omni-3B on WenetSpeech-Chuan. To explore the potential as a foundation model, based on the WenetSpeech-Chuan trained model, we further finetuned an enhanced Paraformer with 1,000 hours of internal high-quality dialectal data.

As shown in Table~\ref{tab:eval_results}, different types of models show performance differences on the Sichuanese dialects test set. First, among all open-source models, FireRedASR demonstrates relatively stable recognition performance across multiple evaluation sets. Notably, FireRedASR-AED achieves an average WER of 15.14\% across all test sets, making it the best-performing open-source model. In contrast, models such as Qwen2.5-omni and kimi-audio exhibit notably higher error rates on the MagicData-Dialogue test set, indicating insufficient adaptability to dialectal speech.

Second, our fine-tuned models show clear performance improvements. After fine-tuning Paraformer and Qwen2.5-omni on WenetSpeech-Chuan, overall performance improves by \textbf{11.7\%} and \textbf{11.02\%} respectively, highlighting the significant effect of WenetSpeech-Chuan in enhancing dialect recognition capability. Furthermore, with an additional 1000 hours of internal data for continued fine-tuning, Paraformer achieves a state-of-the-art average CER of \textbf{13.38\%} across all test sets, confirming that ASR models possess strong transferability and adaptability when trained with high-quality dialectal data.

In summary, our evaluation results clearly demonstrate that, while maintaining Mandarin recognition ability without noticeable degradation (We will show it on our \href{https://github.com/ASLP-lab/WenetSpeech-Chuan}{project page}), WenetSpeech-Chuan substantially enhances models’ capacity to recognize Sichuanese dialects.

\subsection{Speech Synthesis}
To evaluate the effectiveness of WenetSpeech-Chuan, we further train the LLM component of the open-source CosyVoice2 model using the full WenetSpeech-Chuan dataset and denote the system as \textbf{CosyVoice2-WSC}. Moreover, for a fair comparison with the commercial Qwen-TTS, which uses fixed speakers, 
We further supervised fine-tune CosyVoice2-WSC with 100 hours of internal high-quality data from two fixed speakers and denote it as \textbf{CosyVoice2-WSC-SFT}. We then compare CosyVoice2-WSC with several TTS models having dialect support, including Step-Audio-TTS~\cite{huang2025step}, CosyVoice2.0~\cite{du2024cosyvoice}, Llasa-1B~\cite{ye2025llasa} and Qwen-TTS~\footnote{https://github.com/mco2004/qwen-tts}. 

For objective evaluation, we calculate character error rate (CER) using FireRedASR~\cite{cer_xu2025fireredasropensourceindustrialgrademandarin} and speaker similarity (SIM) using WavLM-Large~\cite{sim_Chen2021WavLM}. For subjective evaluation, we adopt Intelligibility-MOS (IMOS), Speaker-MOS (SMOS), and Accent-MOS (AMOS) to assess the comprehensibility, speaker similarity, and accent naturalness of the synthesized speech, respectively. To ensure a more reliable evaluation of AMOS, we invite ten native Sichuanese raters along with ten non-expert listeners. In total, 30 speech samples are evaluated, with 20 drawn from the WSC-Eval-TTS-easy split and 10 from the WSC-Eval-TTS-hard split. All audio prompts are selected from the open-source MagicData test set, covering ten different speakers (five male and five female).

As shown in Table~\ref{tab:comparison}, CosyVoice2-WSC demonstrates competitive performance across both objective and subjective metrics. On the easy split, it reaches a CER of 4.28\%, which is close to  4.13\% achieved by Qwen-TTS, while delivering higher perceptual quality and the best speaker similarity. On the hard split, its CER rises to 8.78\% compared with 7.35\% for Qwen-TTS, yet it still maintains stronger perceptual quality with a SIM above 62\%, showing better robustness in challenging scenarios. Compared with Step-Audio-TTS and the CosyVoice2 baseline, both of which show higher error rates, CosyVoice2-WSC achieves a more favorable balance between accuracy and perceptual quality. After fine-tuning, CosyVoice2-WSC-SFT yields further improvements. On the easy split, it achieves the lowest CER at 4.08\% and the highest SIM at 78.84\%, along with leading MOS-family scores. On the hard split, it lowers the CER to 7.22\% and sustains the best AMOS, demonstrating that fine-tuning enhances both accuracy and perceptual quality. Overall, these results confirm that the WenetSpeech-Chuan dataset provides a solid foundation for building robust and high-quality TTS systems in the Sichuanese dialects.

\vspace{-0.2cm}
\section{CONCLUSION}
In this work, we present WenetSpeech-Chuan, the largest open-source corpus for the Chinese Sichuanese dialects, comprising over 10,000 hours of speech with multi-dimensional annotations. To construct the dataset, we developed the Chuan-Pipeline, a comprehensive data processing toolkit that enabled the creation of this large-scale resource. Furthermore, we established ASR and TTS evaluation benchmarks with manually verified transcriptions to address the lack of standardized testing. Our experiments validate the immense value of this corpus, ASR and TTS models trained on WenetSpeech-Chuan achieve state-of-the-art performance among open-source systems and perform comparably to commercial systems. We anticipate WenetSpeech-Chuan will help advance accelerating innovation across both academic research and industrial applications for dialectal speech processing.

\vfill\pagebreak
\newpage

\bibliographystyle{IEEEbib}
\bibliography{strings,refs}

\end{document}